\documentclass[lettersize,journal]{IEEEtran}
\usepackage{amsmath,amsfonts}
\usepackage{algorithmic}
\usepackage{algorithm}
\usepackage{array}
\usepackage[caption=false,font=normalsize,labelfont=sf,textfont=sf]{subfig}
\usepackage{textcomp}
\usepackage{stfloats}
\usepackage{url}
\usepackage{verbatim}
\usepackage{graphicx}
\usepackage{cite}
\usepackage{booktabs}
\usepackage{multirow}
\usepackage{xcolor}
\usepackage[pagebackref,breaklinks,colorlinks]{hyperref}
\usepackage{xcolor}
\begin{document}

\definecolor{color1}{rgb}{0.51, 0.61, 0.69}
\definecolor{color2}{rgb}{0.57, 0.67, 0.88}
\definecolor{color3}{rgb}{0.96, 0.70, 0.51}
\definecolor{color4}{rgb}{0.99, 0.85, 0.36}
\definecolor{color5}{rgb}{0.67, 0.84, 0.56}
\definecolor{color6}{rgb}{0.49, 0.87, 0.84}
\definecolor{color7}{rgb}{0.96, 0.74, 0.57}
\definecolor{color8}{rgb}{0.00, 0.69, 0.31}

\title{\textbf{\textit{HandEval}}: Taking the First Step Towards Hand  Quality \\ Evaluation in Generated Images}

\author{Zichuan Wang, 
Bo~Peng\textsuperscript{*},~\IEEEmembership{Member,~IEEE,}  
Songlin~Yang,~\IEEEmembership{Student Member,~IEEE,} 

Zhenchen Tang,
and~Jing~Dong,~\IEEEmembership{Senior Member,~IEEE}%
\thanks{\dag indicates corresponding author.}
}

\twocolumn[{
    \renewcommand\twocolumn[1][]{#1}
    \maketitle
        \begin{figure}[H]
        \vspace{-1.0cm}
        \hsize=\textwidth
        \centering
        \includegraphics[width=1.95\linewidth]{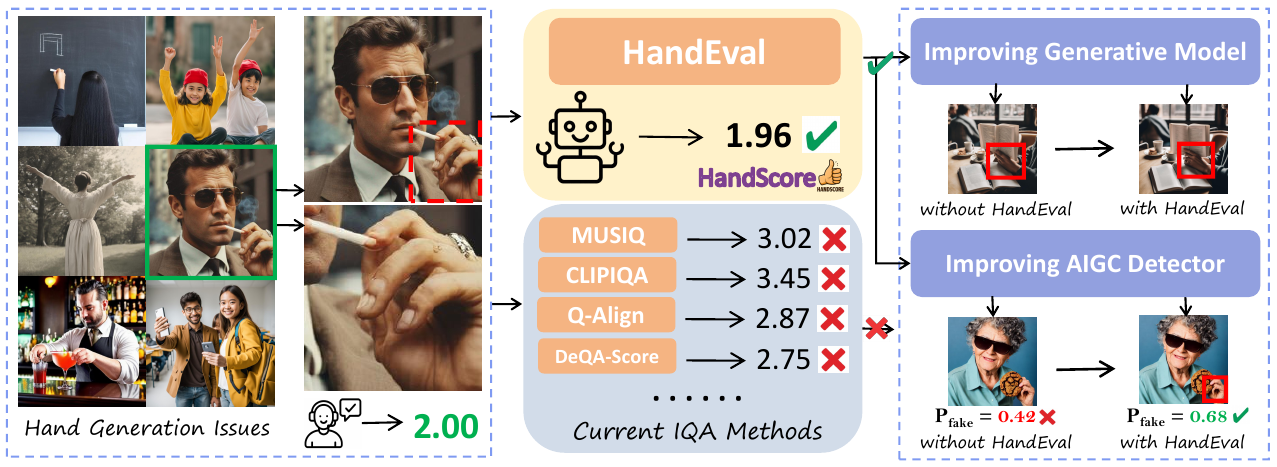}
        \vspace{-0.2cm}
        \caption{Overview of hand quality assessment and its importance for downstream tasks. Hand quality assessment is crucial for ensuring realism in generating human-centric content and detecting AI-generated content (AIGC) where hand artifacts often reveal forgeries. While existing Image Quality Assessment (IQA) methods focus solely on global image quality, they neglect critical hand-specific details and struggle to evaluate hand quality, limiting their application in downstream tasks. Therefore, we propose the first systematic approach to fill the gap of hand-specific quality assessment, achieving high consistency with human ratings as well as improving generation quality and AIGC detection performance.}
        \label{teaser}
        \vspace{-0.2cm}
        \end{figure}
}]

\begin{abstract}
\makeatletter{\renewcommand*{\@makefnmark}{}
\footnotetext{\textsuperscript{*} indicates the corresponding author. Zichuan Wang, Bo Peng, Zhenchen Tang and Jing Dong are with the New Laboratory of Pattern Recognition (NLPR) , Institute of Automation, Chinese Academy of Sciences (CASIA), Beijing, 100190, China. Songlin Yang is with the Hong Kong University of Science and Technology (HKUST), Clear Water Bay, Kowloon, Hong Kong. Zichuan Wang and Zhenchen Tang are also with the School of Artificial Intelligence, University of Chinese Academy of Sciences, Beijing, 100049, China. (E-mail: wangzichuan2024@ia.ac.cn; bo.peng@nlpr.ia.ac.cn; sl.yang888@outlook.com; tangzhenchen2024@ia.ac.cn; jdong@nlpr.ia.ac.cn)
}\makeatother}Although recent text-to-image (T2I) models have significantly improved the overall visual quality of generated images, they still struggle in the generation of accurate details in complex local regions, especially human hands. Generated hands often exhibit structural distortions and unrealistic textures, which can be very noticeable even when the rest of the body is well-generated. However, the quality assessment of hand regions remains largely neglected, limiting downstream task performance like human-centric generation quality optimization and AIGC detection. To address this, we propose the first quality assessment task targeting generated hand regions and showcase its abundant downstream applications. We first introduce the \textit{HandPair} dataset for training hand quality assessment models. It consists of 48k images formed by high- and low-quality hand pairs, enabling low-cost, efficient supervision without manual annotation. Based on it, we develop \textit{HandEval}, a carefully designed hand-specific quality assessment model. It leverages the powerful visual understanding capability of Multimodal Large Language Model (MLLM) and incorporates prior knowledge of hand keypoints, gaining strong perception of hand quality. We further construct a human-annotated test set with hand images from various state-of-the-art (SOTA) T2I models to validate its quality evaluation capability. Results show that \textit{HandEval} aligns better with human judgments than existing SOTA methods. Furthermore, we integrate \textit{HandEval} into image generation and AIGC detection pipelines, prominently enhancing generated hand realism and detection accuracy, respectively, confirming its universal effectiveness in downstream applications. Code and dataset will be available.
\end{abstract}


\section{Introduction}

\IEEEPARstart{R}{ecently}, a photo of “a young boy trapped under the rubble\footnote{\href{https://www.vishvasnews.com/english/ai-check/fact-check-ai-generated-image-of-a-baby-misrepresented-as-survivor-of-nepal-tibet-border-earthquake/}{Fact Check: AI-Generated Image Of a Baby Misrepresented As Survivor Of Nepal-Tibet Border Earthquake}}” circulated widely online. Set against the backdrop of a 6.8-magnitude earthquake in Tibet, it shows a dust-covered boy trapped beneath debris, with a sorrowful expression and an air of helpless isolation, evoking strong emotional responses from viewers. However, this photo is soon revealed to be AI-generated, with a key evidence being the boy’s hand, which appears to have six fingers. This incident not only draws considerable concern over the ethical risks of AI-generated content (AIGC), but also highlights a critical technical flaw: while the overall  quality of generated images has reached a high level, the generation quality of complex local regions, particularly hands, still suffers from significant flaws, reflecting the saying of “The devil is in the details”.  


Human hands are challenging subjects in image generation, as they usually occupy only a small portion of the image and exhibit complex structures. Issues such as extra fingers, missing fingers, anatomical distortions, and abnormal textures frequently occur, even for the most advanced text-to-image (T2I) models like Stable Diffusion (SD)~\cite{sd}, DALL·E~\cite{ramesh2021zero}, and Midjourney (MJ)~\cite{midjourney}. These local defects are hard to eliminate even though the overall quality of generated images continues to improve, which severely hinders the credibility and usability of recent models for generating human-centric images~\cite{realishuman, hand1000, handrefiner, handiffuser}. Moreover, as one of the most intricate and perception-sensitive parts of human body, hands have also become a critical cue in distinguishing between real and fake images, as evident in the aforementioned incident. 
However, using the hands clue for AI generated image detection still lacks research. 
As such, evaluating the “devil” in generated hands is an important topic for both image generation and detection fields.

Therefore, there is an urgent need for a mechanism that can automatically assess the quality of hand regions in generated images. Such a mechanism would support two key tasks: first, it would guide image generation models to improve the quality of hand regions, ensuring AIGC outputs to align more closely with human perception. Second, it could serve as a local detail-aware module to assist in AIGC detection. Unfortunately, existing image quality assessment (IQA) methods are typically designed for global-scale evaluations and perform poorly when it comes to assessing generated hands. The reason lies in the fact that hand structures are usually more complex than other regions, and there is no hands-specific dataset for accomplishing this task.
So there is still a large gap in the quality assessment of hand regions, as shown in Fig.~\ref{teaser}

To fill this gap, we conduct the first systematic study on quality evaluation for human hand regions in generated images, covering three key aspects: dataset curation (\textit{\textbf{HandPair}}), quality assessment methodology (\textit{\textbf{HandEval}}), and practical applications in improving generation and detection tasks. 

\textbf{(a) Dataset Curation}: We begin by constructing \textit{\textbf{HandPair}}, the first dataset dedicated to training hand quality evaluation models (\textit{Quality evaluation} and \textit{quality assessment} are used interchangeably in this paper, as they denote the same concept without causing confusion). Unlike traditional approaches that rely on labor-intensive manual scoring, \textit{HandPair} adopts a positive–negative pairing strategy: by applying HandRefiner~\cite{handrefiner} to modify the hand region in real images, we automatically collect paired examples of high-quality and low-quality hands. This strategy significantly improves the efficiency of dataset construction and reduces annotation costs, and also ensures paired images differ only in quality while sharing consistent background and pose, enabling the model to focus on learning hand quality specific features. The dataset contains a total of 48k hand images (24k high-quality and 24k low-quality), providing a solid foundation for hand quality assessment. 

\textbf{(b) Quality Assessment}: Based on \textit{HandPair} dataset, we propose \textbf{\textit{HandEval}}, the first model specifically designed to evaluate the generation quality of hand regions. It incorporates hand keypoint priors, acknowledging that hand quality assessment heavily relies on structural integrity, and leverages the strong visual understanding capability of MLLMs to accurately perceive hand quality. To evaluate its effectiveness, we further construct an independent test set containing images with human hands generated by various SOTA T2I models, each annotated by human experts to obtain ground truth quality scores for the hand region. Experimental results demonstrate that \textit{HandEval} significantly outperforms existing IQA methods and achieves SOTA performance on this task.

\textbf{(c) Applications}: Furthermore, we explore the potential of \textit{HandEval} in \textbf{downstream applications} that benefit from its ability to assess hand region quality. In image generation, we integrate \textit{HandEval} into the Concept Slider~\cite{concept} framework, using its scores as supervisory signals to guide the model toward generating higher-quality hands. In AIGC detection, we use \textit{HandEval} as a plug-and-play local defect signal, which is fused with the outputs of existing detectors to improve detection accuracy. Our experiments show that \textit{HandEval} consistently enhances performance in both tasks, validating its practical utility and generalizability in real-world scenarios.

In summary, our contributions are as follows:
\begin{itemize}
\item We are the first to formally introduce and systematically study the problem of quality assessment for hand regions in generated images.
\item We construct \textit{HandPair}, the first dataset tailored for training hand quality evaluation models, and develop \textit{HandEval}, the first high-performance hand quality evaluator.
\item We conduct evaluation experiments and demonstrate the strong hand quality assessment capability of \textit{HandEval}. We also incorporate it into downstream tasks, showing its ability to improve both hand generation quality and AIGC detection performance, highlighting its strong practical value and extensibility. 
\end{itemize}


\section{Related Works}
\subsection{Non-MLLM-based Image Quality Assessment}
Image Quality Assessment (IQA) is a fundamental research direction in computer vision, aiming to simulate human perception of visual quality. IQA methods are typically categorized into three types: full-reference (FR), no-reference (NR), and reduced-reference (RR). Traditional FR methods, such as PSNR and SSIM~\cite{ssim}, assess image quality by comparing a distorted image to its high-quality reference. NR methods attempt to estimate perceived quality without any reference images, with representative approaches including BRISQUE~\cite{brisque}, NIQE~\cite{niqe}, and PIQE~\cite{piqe}.
Recently, with the rapid development of deep learning, learning-based IQA methods have become increasingly prominent. 
Methods such as LPIPS~\cite{lpips} and DISTS~\cite{dists} have improved alignment with human perception under full-reference settings. In the no-reference scenario, models like PaQ-2-PiQ~\cite{paq2piq}, MANIQA~\cite{maniqa}, and HyperIQA~\cite{hyperiqa} utilize CNN or Vision Transformer (ViT) architectures to extract features from entire images and regress quality scores. 
Additional strategies, including multi-task learning, spatial attention mechanisms, and contrastive paradigms~\cite{contrique, clipiqa, clipagiqa}, have also been explored to enhance the joint modeling of content and quality.

However, conventional IQA methods are primarily designed for natural image quality assessment tasks and may not generalize well to AI generated images. The former focuses on measuring low-level distortions in images, while the latter emphasizes evaluating structural plausibility and semantic realism. In recent years, researchers have proposed various metrics specifically for evaluating generated images. 
Metrics like Fréchet Inception Distance (FID)~\cite{fid} and Inception Score (IS)~\cite{is} are widely used to evaluate the visual quality of generated images, while CLIPScore~\cite{clipscore} evaluates the image–text alignment quality.
More recent approaches, such as ImageReward~\cite{imagereward}, PickScore~\cite{pickscore}, and HPS~\cite{hpsv2}, incorporate human preference to assess the overall plausibility and relevance of generated images. Meanwhile, a growing number of benchmark datasets~\cite{agiqa3k, agin, evalmuse} targeting generated image quality have also been introduced to support robust evaluation.

\subsection{MLLM-based Image Quality Assessment}
In recent years, Multimodal Large Language Models (MLLMs), such as BLIP-2~\cite{blip2}, LLaVA~\cite{llava}, and QwenVL~\cite{qwenvl2}, have demonstrated remarkable performance in tasks like image-text understanding and visual question answering. With strong generalization capabilities in capturing visual details and semantic reasoning, MLLMs have opened up new possibilities for visual quality assessment. Among them, Q-Align~\cite{qalign} introduces a novel quality-aware paradigm by fine-tuning MLLMs under supervised instructions, enabling them to assess image quality based on user prompts. Building on this structure, Deqa-Score\cite{deqa_score} adds a KL divergence loss between predicted and target score distributions to address the limitation of MLLMs in producing continuous quality scores. Similarly, models such as Decipt-IQA~\cite{depictqa}, Compare2Score~\cite{compare2score}, and EvalAlign~\cite{evalalign} leverage the next-token prediction nature of MLLMs and design various instruction formats, ranging from direct quality estimation, pairwise comparison, to quality reasoning, to generate more diverse outputs and enhance both performance and interpretability in quality assessment. This paradigm has proven more effective in assessing image quality than Non-MLLM-based IQA methods, so we follow it to evaluate the quality of hand regions.

Despite the success of above methods, most of them adopt a global image modeling paradigm, operating at coarse granularity and being sensitive to localized distortions in semantically and structurally complex regions such as hands. This issue is particularly prominent in generated images, where models often generate visually consistent results at the global level while exhibiting subtle structural anomalies in local regions. Existing IQA models struggle to capture these fine-grained quality degradations, posing a significant obstacle to further enhancement of generation quality.

\subsection{Localized Quality Assessment }
To address the limitations of coarse-grained global quality assessment, some studies have begun to focus on localized image quality assessment (Localized IQA) for key regions. Among these, face quality assessment is the most extensively studied, with widespread applications in face recognition, and video surveillance. Representative methods such as FaceQnet~\cite{faceqnet}, SER-FIQ~\cite{ser-fiq} and CRL-GFIQA~\cite{crl-gfiqa} typically model factors like facial clarity, structural integrity, and expression consistency, forming a relatively mature line of research. In addition, FaceScore~\cite{facescore} proposes a new face quality metric tailored for generated images by leveraging inpainting-based (win, loss) face pairs and human-aligned preference learning. However, it does not exploit information unique to facial regions, and its contrastive learning framework only enables relative comparisons of facial quality rather than producing accurate scores, limiting its applicability in downstream tasks.

Compared to the face, the human hand is structurally more complex, exhibiting greater pose variability and more complex hand-object interactions, and is more prone to distortion in generated images, yet lacks dedicated quality assessment mechanisms. To date, there are no established models or publicly available training datasets specifically targeting hand region quality. Existing IQA methods also struggle to reliably evaluate the hand area. In this work, we present the first quality assessment task specifically designed for the hand region in generated images, and construct a dedicated dataset and evaluation model, filling the gap in this area.

\begin{figure*}[htbp]
    \centering
    \includegraphics[width=0.98\linewidth]{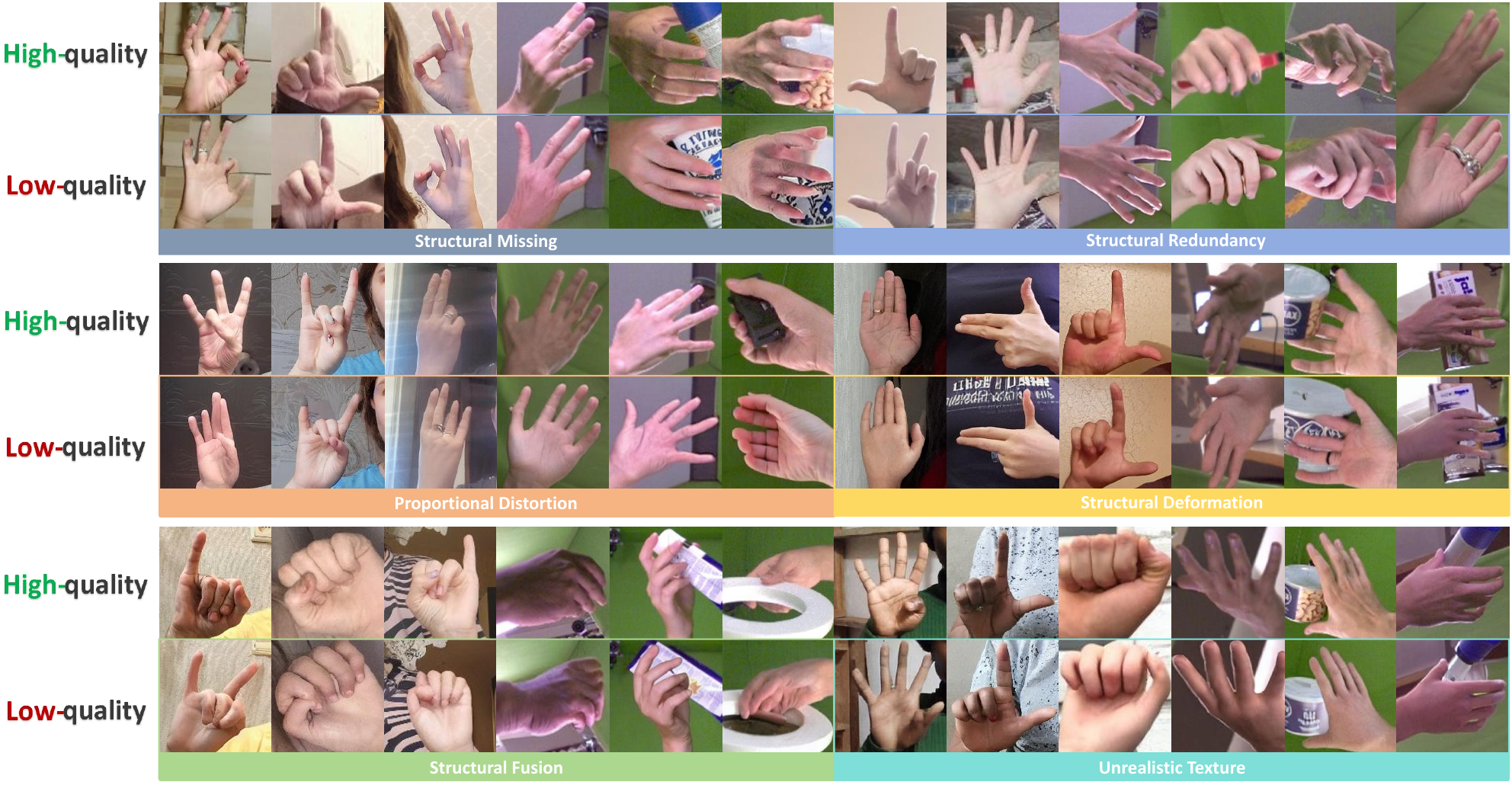}
    \caption{Example image pairs from the \textit{HandPair} dataset. Each pair consists of a high-quality hand image and its corresponding low-quality version generated using hand-specific inpainting methods. The low-quality images exhibit typical visual defects in generated hands, which are categorized into six types: \colorbox{color1}{structural missing}, \colorbox{color2}{structural redundancy}, \colorbox{color3}{proportional distortion}, \colorbox{color4}{structural deformation}, \colorbox{color5}{structural fusion} and \colorbox{color6}{unrealistic texture}.}
    \label{fig:dataset}
\end{figure*}
\section{\textit{HandPair}: The Hand Quality Assessment Training Dataset}
In this section, we introduce \textit{HandPair}, the first dataset specifically constructed for training hand quality assessment models. Unlike traditional quality assessment datasets that rely heavily on extensive manual annotations, \textit{HandPair} adopts a positive–negative pair structure, where high-quality hand images are sourced from existing datasets of real hands, and low-quality counterparts are automatically generated using hand-specific degradation techniques. This approach significantly reduces annotation costs and improves the efficiency of dataset construction. Furthermore, it ensures that the real and generated hand images share a more similar data distribution, with consistent background and general hand pose, differing only in terms of hand quality. This alignment helps the model to focus more effectively on learning quality-related features. Some examples of \textit{HandPair} are shown in Fig.~\ref{fig:dataset}. The following subsections detail the dataset curation pipeline and provide a comprehensive data analysis.

\subsection{Dataset Construction}
\subsubsection{Raw Image Collection}
We randomly selected images from two publicly available hand-related image datasets, HAGRID~\cite{hagrid} and FreiHAND~\cite{freihand}, as initial source images. HAGRID is a hand gesture recognition dataset, and FreiHAND is a hand pose estimation dataset. Both contain diverse hand poses, skin tones, angles, and background environments. We sampled 20,000 images from each dataset, totaling 40,000 images, serving as the foundation of high-quality hand images.

\subsubsection{Automated Degradation Generation}
To construct low-quality hand images, we innovatively repurposed HandRefiner~\cite{handrefiner} in a reverse manner. It is originally designed to correct malformed hand regions in generated images through conditional diffusion and inpainting. But in its pipeline, the quality of the generated hands can be significantly influenced by adjusting a control strength parameter. Based on it, we defined three levels of control strength: 0.4, 0.55 and 0.7, to deliberately synthesize hand images with different types of degradation. Specifically, a control strength of 0.4 produces images with severe structural distortions but relatively realistic textures, while 0.7 results in more structurally accurate hands with degraded textures. The 0.55 setting yields an intermediate balance between the two. In addition, given that our original high-quality images already exhibit a wide variety of hand poses, this strategy effectively ensures diversity in the generated low-quality hand images.

\subsubsection{Hand Cropping and Filtering}
After collecting both high- and low-quality hand images, we employed MediaPipe~\cite{mediapipe} to detect and locate hand keypoints in each image. Based on the detected hand bounding-boxes, cropping was performed centered on these bounding-boxes, with the crop box expanded to 1.2 times the original size to retain appropriate context. To ensure dataset quality, images with hand regions that were too small (height or width less than 80 pixels) were removed. High-quality hand images were then paired one-to-one with corresponding low-quality images, ensuring balance between positive and negative samples.

\subsection{Dataset Analysis}
\subsubsection{Dataset Scale}
Through the aforementioned data construction process, the \textit{HandPair} dataset comprises 24,066 pairs of hand images (a total of 48k images). Each pair consists of one high-quality hand image and one low-quality counterpart. The dataset includes both empty gestured hands and hands interacting with objects, effectively simulating real-world hand usage scenarios. The proportions of these two types are $65.14\%$ and $34.86\%$, respectively.

\subsubsection{Hand Pose Statistics}
We use MediaPipe to extract hand keypoints from high-quality images. These same keypoints are also used to generate corresponding low-quality images using HandRefiner, so they have similar hand poses to high-quality ones. Based on the keypoints, we conduct a diversity analysis in two aspects: finger flexion angles and palm orientation.

\begin{figure*}[htbp]
    \centering
    \includegraphics[width=1.0\linewidth]{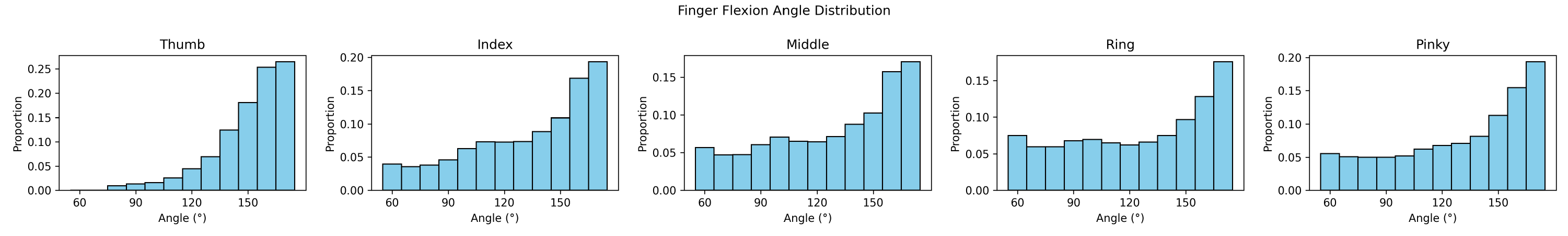}
    \caption{Distribution of finger flexion angles computed from PIP joints. A smaller angle indicates a more bent finger. A higher proportion of large-angle samples (~150° and above) is observed,  which can be attributed to frequent hand-object interaction scenarios where fingers tend to be more extended in our dataset. Overall, the distributions confirm diverse finger poses across the dataset, with the thumb exhibiting consistently larger angles due to its anatomical characteristics.}
    \label{fig:finger_flexion_hist}
\end{figure*}

First, we use the PIP (Proximal Interphalangeal) joint angle of each finger to represent its degree of flexion~\cite{dipietro2008survey}, which is located in the middle segment of each finger and is responsible for the bending and extending of the finger. The distribution results are shown in Fig.~\ref{fig:finger_flexion_hist}, where a smaller angle indicates a more bent finger. As observed, the flexion angles of the four non-thumb fingers are all above 60°, and the thumb angles are typically above 80°, which aligns with human anatomy. A relatively high proportion of samples fall within the large-angle range (approximately 150° and above), which can be attributed to the presence of hand-object interaction scenarios in the dataset. In such cases, fingers tend to be more extended. Excluding this factor, the distribution of flexion angles across the four fingers remains relatively uniform, indicating that the collected hand images exhibit a high degree of pose diversity. The thumb consistently shows higher angle values, which is expected, as thumbs are often extended in everyday gestures.

\begin{figure}[htbp]
    \centering
    \includegraphics[width=1.0\linewidth]{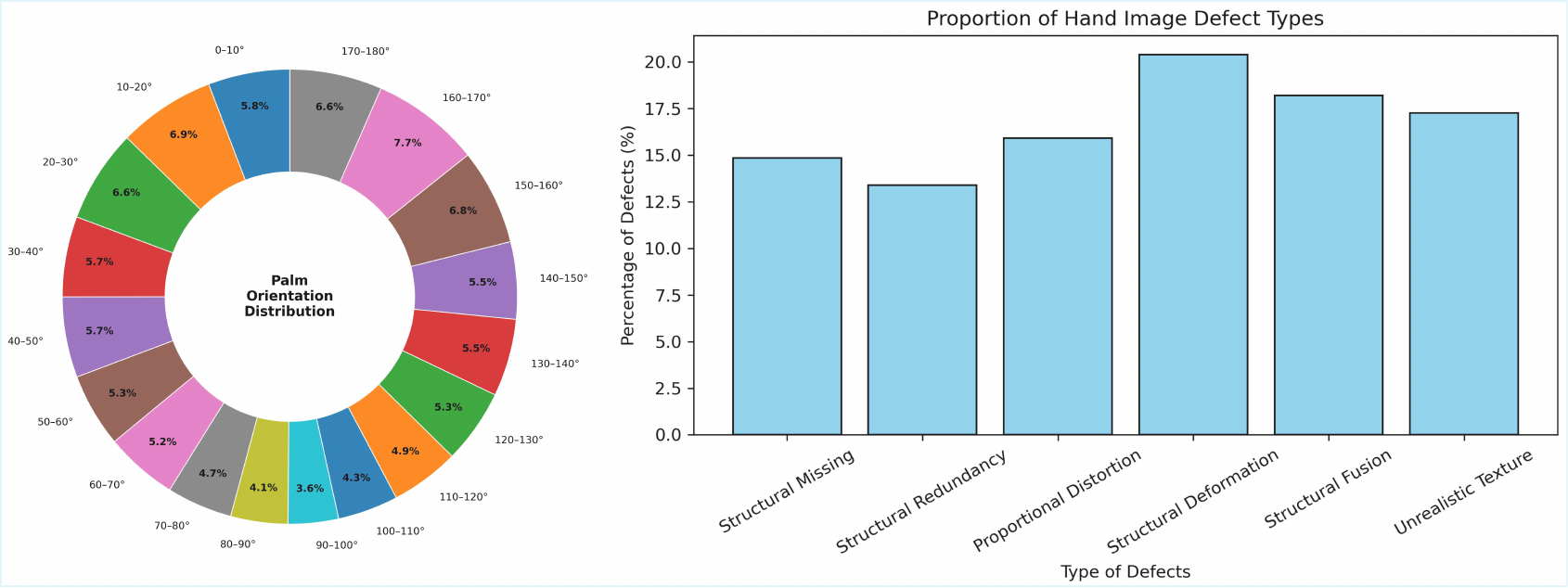}
    \caption{Palm orientation distribution (left) and proportion of each hand image defect type (right). The palm orientation distribution shows a relatively uniform angular coverage from 0° to 180°, indicating diverse viewpoints without significant bias. The sample counts across six defect categories are relatively balanced, ensuring diversity and representativeness of the generated low-quality hand images.}
    \label{fig:merge_pie_defect}
\end{figure}

Then, we analyze palm orientation by computing the angle between the palm normal vector and the image Z-axis (i.e., the camera viewing direction). The distribution is illustrated on the left of Fig.~\ref{fig:merge_pie_defect}. The results show a relatively uniform distribution across the full range of 0°–180°, without significant angular bias or missing regions. This indicates that our dataset covers a wide range of hand orientations, from palm facing forward to backward, and avoids viewpoint imbalance.

\subsubsection{Defect Type Statistics}
We further categorize the low-quality hand images based on common visual defects. These defects are grouped into six categories:
\begin{itemize}
\item[-] Structural Missing: missing fingers or incomplete anatomical regions;
\item[-] Structural Redundancy: extra fingers or repeated joints;
\item[-] Proportional Distortion: abnormal proportions, such as unusual finger lengths or finger-palm size ratios;
\item[-] Structural Deformation: all hand components exist but exhibit unnatural twisting or bending;
\item[-] Structural Fusion: different parts of the hand are merged or entangled without clear anatomical boundaries;
\item[-] Unrealistic Texture: texture artifacts such as blurred surfaces, or unnatural material appearance.
\end{itemize}

We calculate the proportions of these six defects in the dataset. To save time, we randomly sample 4k images from the low-quality hand images for statistical analysis. The right of Fig.~\ref{fig:merge_pie_defect} shows the proportion of each defect. Since each image may contain more than one defect, the sum of the proportions of these six defects exceeds 1. As observed, the sample counts for each category are relatively balanced, without any extreme dominance or scarcity. This confirms that our degraded image generation strategy produced a diverse and representative set of hand defects, which helps ensure the robustness of quality assessment models across various defect scenarios.

\section{\textit{HandEval}: The Hand Quality Assessment Model}
\begin{figure*}[htbp]
    \centering
    \includegraphics[width=0.95\linewidth]{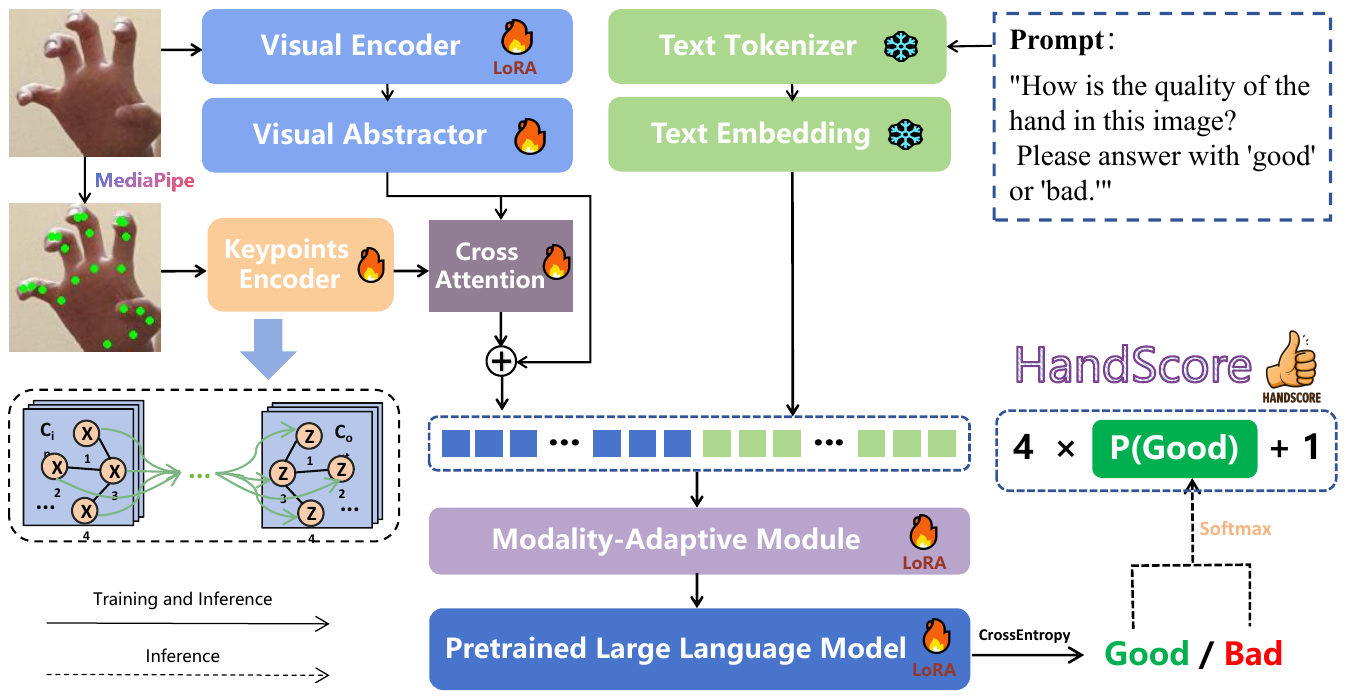}
    \caption{The overall architecture of \textit{HandEval}. It incorporates hand keypoint priors to enhance quality assessment. Hand images are processed through a visual encoder, while structural priors in the form of hand keypoints are encoded via a GCN-based keypoint encoder to capture spatial and structural information. These visual and keypoint features are then fused through a cross-attention mechanism. Finally, guided by a text-based prompt, a pretrained language model performs quality assessment and outputs the final HandScore.}
    \label{fig:handeval}
\end{figure*}

Most existing quality assessment models are built upon global visual quality, limiting their ability to accurately perceive localized regions such as hands, as shown in Table~\ref{tab:eval} of the experiment section.  To address it, we propose \textit{HandEval}, the first model specifically designed for hand quality assessment. Our model is constructed on top of the MLLM framework mPLUG-Owl2~\cite{mplug-owl2}, which has demonstrated excellent image quality perception capabilities in many works~\cite{qalign, deqa_score, qbench, qinstruct}. Based on this foundation, we incorporate our \textit{HandPair} dataset and design hand-specific natural language instructions for supervised fine-tuning. Furthermore, we introduce prior knowledge from hand keypoints to enhance the model's structural awareness of hand regions. The overall architecture is illustrated in Fig.~\ref{fig:handeval}. By integrating visual images, hand keypoint information, and natural language instructions, \textit{HandEval} can effectively guide the model to focus on hand-region quality, thus overcoming the local detail perception limitations of conventional quality assessment models.

\subsection{Hand Quality Perception Enhanced by Keypoint Priors}
Conventional image quality assessment mainly relies on global features extracted by vision encoders, which are inadequate for modeling the highly structured and complex nature of human hands. To address this limitation, we introduce a keypoint prior enhancement mechanism, leveraging structural guidance to improve accuracy in hand quality evaluation.

The model input consists of three components: hand image $I$, keypoints $K$, and natural language instruction $T$. Hand keypoints $K$ are extracted via MediaPipe. To leverage the prior knowledge of hand keypoints, we design a keypoint encoder based on Graph Convolutional Network (GCN)~\cite{GCN}, which is naturally well-suited for the graph-structured nature of hand joints. The GCN maps $K$ into high-dimensional structural embeddings that preserve both geometric and relational information. Then these graph-enhanced embeddings are used as queries in a cross-attention module, interacting with visual features extracted from the image encoder. The fusion process is formulated as:
\begin{equation}
C(I, K) = \text{Attention}(Q = E_k(K),\; K = E_i(I),\; V = E_i(I)),
\end{equation}
\begin{equation}
F(I, K) = \text{LayerNorm}(C(I, K) + E_i(I)),
\end{equation}
where $E_k(\cdot)$ and $E_i(\cdot)$ denote the keypoints encoder and image encoder respectively, and $F(I, K)$ represents the fused visual feature. This simple yet effective fusion mechanism explicitly models hand-specific quality features, which are subsequently integrated with textual features for hand quality evaluation.

\subsection{Fine-tuning Process and Training Objective}
The model is built upon the MLLM framework mPLUG-Owl2, denoted as $f_\theta(\cdot)$, where $\theta$ represents model parameters. The input is a triplet $(I, K, T)$. $I$ and $K$ are used to generate the fused visual features, as described in the previous section, and $T$ is passed through a text encoder to obtain the corresponding textual features. These two modalities are then jointly fed into the Modality-Adaptive Module for feature alignment. The resulting aligned representation is subsequently passed into a pretrained large language model, which produces the final output, a token sequence $Y = \{y_1, y_2, \dots, y_n\}$, where $y_i$ denotes the $i$-th generated token. The target vocabulary is defined as $C = \{\texttt{good}, \texttt{bad}\}$, corresponding to the high- and low-quality hand images in our dataset.

During training, the model maximizes the conditional probability:
\begin{equation}
\max_{\theta} P(Y \mid I, K, T; \theta)
\end{equation}
using the standard cross-entropy loss:
\begin{equation}
\mathcal{L}(\theta) = - \sum_{i=1}^{n} \log P(y_i \mid y_{<i}, I, K, T; \theta).
\end{equation}
Since the task is word-level classification, the generated sequence is designed to be short (i.e., $n=1$), and training focuses on correctly predicting the final quality label.

The instruction template is designed as:
\begin{quote}
\texttt{Prompt: "How is the quality of the hand in this image? Please answer with 'good' or 'bad'.<image>"}
\end{quote}
which explicitly guides the model to focus on hand quality.

\subsection{Quality Scoring}

During inference, the model generates prediction probabilities for both candidate tokens:
\begin{align}
p_\text{good} &= P(y = \texttt{"good"} \mid I, K, T; \theta), \\
p_\text{bad} &= P(y = \texttt{"bad"} \mid I, K, T; \theta).
\end{align}

We apply softmax between the two tokens and define the final hand quality score in the range of 1 to 5 as:
\begin{equation}
S_\text{hand} = 4 \cdot \frac{e^{z_\text{good}}}{e^{z_\text{good}} + e^{z_\text{bad}}} + 1,
\end{equation}
where $z_\text{good}$ and $z_\text{bad}$ are the logit outputs. This score directly reflects the model’s confidence in assessing hand quality and can serve as a quantitative metric in AIGC image optimization and AIGC detection tasks.

\section{Downstream Applications}
In the previous section, we introduce the \textit{HandEval} model, which is capable of performing targeted quality assessment of hand regions. Moreover, the utility of \textit{HandEval} extends beyond quality assessment. Its powerful perceptual ability in assessing hand quality opens up potential use in generation and AIGC detection tasks. In this section, we provide a detailed description of these two applications respectively. 

\subsection{HandEval-Guided Quality Optimization}
\begin{figure}[htbp]
    \centering
    \includegraphics[width=1.0\linewidth]{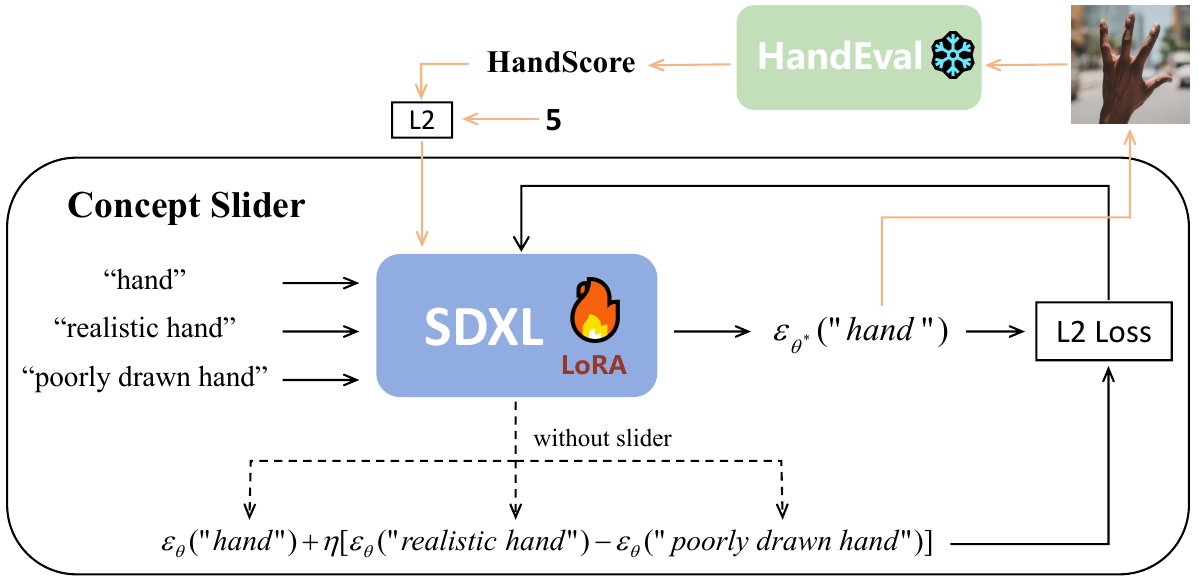}
    \caption{The framework of \textit{HandEval} incorporated into Concept Slider. The structure inside the rounded rectangle represents the Concept Slider. The {\color{color7}{$\leftarrow$}} indicates how our \textit{HandEval} is integrated into it.}
    \label{fig:generation}
\end{figure}

In this section, we investigate how the \textit{HandEval} can be incorporated into the training pipeline of diffusion models to explicitly guide the enhancement of hand generation quality. Specifically, we build upon the Concept Slider framework~\cite{concept}, embedding the \textit{HandEval} output as an external quality signal to supervise and steer the generation process toward high quality hand. The framework is shown in Fig.~\ref{fig:generation}

Concept Slider is a technique for controlling the output of image generation. It works by using a pair of semantically opposite prompts (e.g., “realistic hand” and “poorly drawn hand”) to guide the model in a specific direction. By learning the difference between these two concepts, the model can enhance the expression of the desired concept (e.g., “realistic hand”) while suppressing the undesired one (“poorly drawn hand”) during image generation.
However, this approach remains essentially language-driven, lacking direct perception and feedback on the generated visual details. To address this limitation, we extend the framework by incorporating \textit{HandEval}-based quality feedback into the training process, allowing for explicit supervision to optimize hand generation quality.
Concretely, we first generate image samples during the training of Concept Slider, and extract the hand regions from them using MediaPipe. Then these samples are fed into the pretrained hand quality evaluator \textit{HandEval}, which outputs a quality score $q \in [1, 5]$, representing the visual quality level of the hand region in the generated image.

To incorporate this feedback into training, we define an auxiliary loss term as:

\begin{equation}
\mathcal{L}_{\text{quality}} = \frac{1}{N} \sum_{i=1}^{N} (q_i - 5)^2
\end{equation}
where $N$ denotes the number of samples. This loss penalizes deviations from ideal hand quality, with lower scores incurring higher penalties. The total training objective becomes:

\begin{equation}
\mathcal{L}_{\text{total}} = \mathcal{L}_{\text{denoise}} + \lambda \cdot \mathcal{L}_{\text{quality}}
\end{equation}
where $\mathcal{L}_{\text{denoise}}$ is the original noise prediction loss in the diffusion model, and $\lambda$ is a hyperparameter that balances the influence of quality supervision, which is set to 0.1 in our experiment. 

\subsection{HandEval-Enhanced AIGC Detection}
In this section, we explore the potential of \textit{HandEval} for a different task: detecting AI generated images. The key motivation for this approach is that as AIGC images increasingly approximate real images in terms of global visual quality, localized regions, e.g. human hands, emerge as crucial indicators for identifying authenticity. Based on this motivation, we design a plug-and-play AIGC detection enhancement module by embedding \textit{HandEval} into existing AIGC detection pipelines. This module leverages the hand quality scores provided by \textit{HandEval} as an auxiliary signal to complement the original detector’s prediction, thereby improving the accuracy of the detection system. The method is detailed below. 

Given an input image $\mathbf{I}$, a pretrained AIGC detector outputs a probability indicating whether the image is fake:
\begin{equation}
P_{\text{detector}} = f_{\theta}(\mathbf{I}) \in [0, 1]
\end{equation}
where $f_{\theta}(\cdot)$ denotes the parameters of the pretrained detector.

Simultaneously, the same image is passed to the \textit{HandEval} model $h_{\phi}(\cdot)$, which outputs a hand quality score:
\begin{equation}
S_{\text{hand}} = h_{\phi}(\mathbf{I}) \in [1, 5]
\end{equation}
where $S_{\text{hand}}$ is a scalar indicating the perceptual quality of the hand region. A score closer to 1 implies low-quality (potentially fake) hands, whereas a score closer to 5 indicates natural, high-quality hand image.

We then combine the two signals using a weighted fusion function:

\[
P_{\text{fused}} = 
(1-\alpha) \cdot P_{\text{detector}} + \alpha \cdot S
\]
Here, $S = (5 - S_{\text{hand}})/4 \in [0, 1]$, where a higher value indicates lower hand quality. $\alpha$ is the weighting parameter.  

The core idea behind this fusion mechanism is intuitive: when the hand quality is poor, the likelihood that the hand region is forged is relatively high. Therefore, we enhance the detection probability by weighting it according to the hand quality. Conversely, when the hand quality is high, it indicates a lower forgery risk, which is typically challenging to generate accurately. We correspondingly reduce the detection probability, reflecting the increased reliability of the image.

\section{Experiments}
\subsection{Performance of Hand Quality Assessment}
\begin{figure}[htbp]
    \centering
    \includegraphics[width=1.0\linewidth]{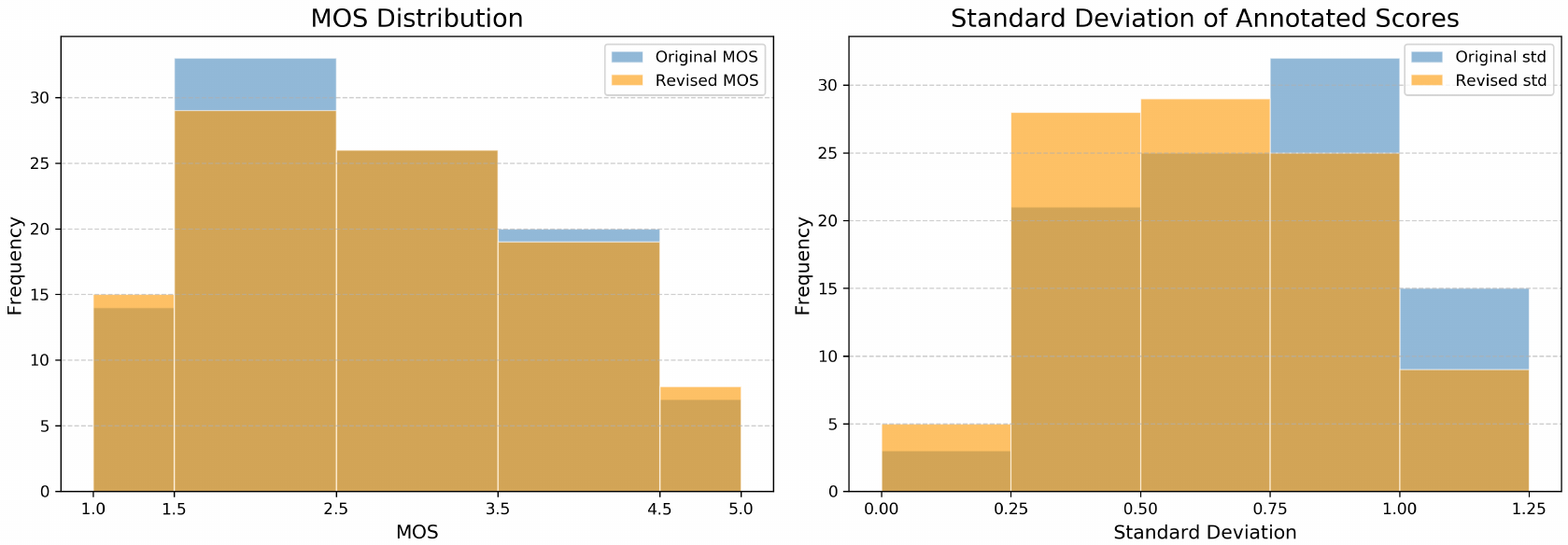}
    \caption{Distribution of Original and Revised MOS and their standard deviations. We use the 2$\sigma$ criterion to remove outlier scores, leading to reduced score dispersion and a more balanced distribution across quality levels.}
    \label{fig:mos}
\end{figure}

Since our model is not trained using explicit quality scores, its actual quality assessment capability needs to be verified. To address this, we construct an independent test set consisting of 100 images. These images are generated by 10 state-of-the-art T2I models, including SDXL, SD 3.5, DALL·E 3, Midjourney v6.1, Doubao, Wanxiang 2.1, FLUX.1, Pixart-$\alpha$, Playground 2.5 and Cogview 3 plus. Each image is rated based on hand quality by eight human experts, taking into account both hand structure and texture. To eliminate outlier scores with significant deviation, we apply the 2$\sigma$ criterion in statistics. The distribution of Mean Opinion Scores (MOS) and their standard deviations before and after this correction are shown in Fig.~\ref{fig:mos}. As can be seen, after filtering, the distribution of the standard deviations is more concentrated toward the lower-variance end, indicating greater consistency among annotators' scores. Meanwhile, the subjective scores are more evenly distributed across different quality levels, ensuring the representativeness and coverage of the test dataset. We then conduct evaluation experiments on this dataset, comparing our method with several existing quality assessment approaches, using three widely adopted metrics, 
PLCC (Pearson Linear Correlation Coefficient), SRCC (Spearman Rank-Order Correlation Coefficient), and KRCC (Kendall Rank Correlation Coefficient) as evaluation criteria.  

\subsubsection{Comparasion with other IQA methods} 
\begin{table*}[htbp]
  \centering
  \caption{Comparison of different IQA methods for hand quality assessment. The “Images Level” denotes the correlation between predicted score and MOS for each image, while the “T2I Models Level” refers to the correlation between the average predicted score and average MOS across different T2I models. “+" indicates methods that can be trained using our high- and low-quality samples and further fine-tuned on our dataset. The best result is in \textbf{bold}, and the second-best is \underline{underlined}.}
  
  \label{tab:eval}
  \resizebox{0.72\linewidth}{!}{
  \begin{tabular}{l|ccc|ccc}
    \toprule
    \multirow{2}{*}{Method} 
    & \multicolumn{3}{c|}{Images Level} 
    & \multicolumn{3}{c}{T2I Models Level} \\
    \cmidrule(lr){2-4} \cmidrule(lr){5-7} 
    & PLCC & SRCC & KRCC & PLCC & SRCC & KRCC \\
    \midrule
    BRISQUE~\cite{brisque} & 0.0059 & 0.0255 & 0.0280 & 0.0564 & 0.0545 & 0.0182\\
    NIQE~\cite{niqe} & 0.1465 & 0.1203 & 0.0912 & 0.3193 & 0.3000 & 0.2364 \\
    UNIQUE~\cite{unique} & 0.1869 & 0.1731 & 0.1174 & 0.4215 & 0.3909 & 0.2364 \\
    CNNIQA~\cite{cnniqa} & 0.3062 & 0.3401 & 0.2234 & 0.4956 & 0.5273 & 0.4182 \\
    DBCNN~\cite{dbcnn} & 0.3575 & 0.3696 & 0.2747 & 0.5372 & 0.5455 & 0.4182 \\
    HyperIQA~\cite{hyperiqa} & 0.4867 & 0.4892 & 0.3604 & 0.6737 & 0.5818 & 0.4909 \\
    ManIQA~\cite{maniqa} & 0.4893 & 0.4887 & 0.3485 & 0.6568 & 0.6182 & 0.4545 \\
    MUSIQ~\cite{musiq} & 0.5212 & 0.5319 & 0.3830 & 0.6834 & 0.6273 & 0.4182 \\
    CLIPIQA~\cite{clipiqa} & 0.4037 & 0.4232 & 0.2921 & 0.6074 & 0.5727 & 0.3818 \\
    CLIPIQA+ & 0.4973 & 0.5470 & 0.3984 & 0.6899 & 0.6374 & 0.5021 \\
    PickScore~\cite{pickscore} & 0.0002 & 0.0026 & 0.0059 & -0.2345 & -0.2545 & -0.2364 \\
    PickScore+ & 0.1667 & 0.1659 & 0.1132 & 0.3753 & 0.3214 & 0.2589 \\
    ImageReward~\cite{imagereward} & 0.2821 & 0.2794 & 0.1925 & 0.1851 & 0.2273 & 0.2000 \\
    ImageReward+ & 0.3292 & 0.3105 & 0.2136 & 0.4754 & 0.4869 & 0.4430  \\
    HPSv2~\cite{hpsv2} & 0.0106 & 0.0590 & 0.0246 & -0.2877 & -0.3722 & -0.2364 \\
    HPSv2+ & 0.1933 & 0.1450 & 0.1052 & 0.4027 & 0.3051 & 0.3325 \\
    Q-Align~\cite{qalign} & 0.5214 & 0.5287 & 0.3730 & 0.6949 & \underline{0.7455} & \underline{0.6364} \\
    DeQA-Score~\cite{deqa_score}  & \underline{0.5511} & \underline{0.5708} & \underline{0.4111} & \underline{0.7182} & 0.7091 & 0.6000 \\
    \textbf{HandEval(Ours)} & {\textbf{0.6693}} & {\textbf{0.6567}} & {\textbf{0.4973}} & {\textbf{0.8637}} & {\textbf{0.9093}} & {\textbf{0.7820}} \\
    \bottomrule
  \end{tabular}}
\end{table*}

\begin{figure*}[!htbp]
    \centering
    \includegraphics[width=0.98\linewidth]{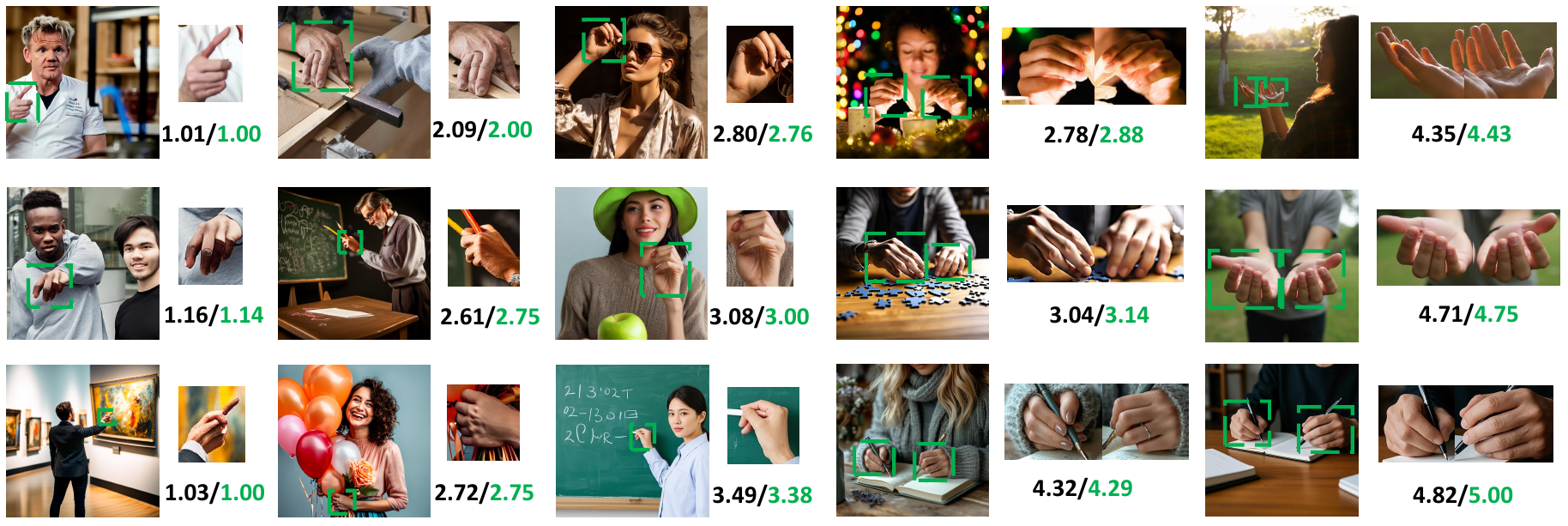}
    \caption{Visual examples of evaluation results. The \textbf{black} numbers represent our model’s predicted scores, and the {\color
    {color8}green} numbers represent the ground truth labels given by human annotators. The score of an image that has multiple hands is the average over all hand scores.}
    \label{fig:eval_example}
\end{figure*}

Table~\ref{tab:eval} reports the evaluation results of our proposed method and existing IQA approaches across two dimensions: the images level, which measures the correlation between predicted scores and the ground truth MOS for each image, and the models level, which measures the correlation between the average predicted scores and the average ground truth MOS for each T2I model. Here, the models level metrics reflect the ability of evaluating a generation model's overall hand generation quality. For most methods, we use their publicly released pretrained weights, as these models require explicit quality scores for supervised training and cannot be directly trained on our paired high- and low-quality dataset. Other methods (CLIPIQA~\cite{clipiqa}, PickScore~\cite{pickscore}, ImageReward~\cite{imagereward}, and HPSv2~\cite{hpsv2}) are compatible with our dataset structure and support training with relative quality preferences. Therefore, we also report their fine-tuned results on our dataset (denoted by a “+”).

At the image level, \textit{HandEval} outperforms all other methods across all three metrics, demonstrating its superior ability to capture perceptual features relevant to hand quality and more closely align with human evaluation standards. At the model level, \textit{HandEval} again achieves the best results, reflecting its capability not only to predict image-level quality but also to accurately distinguish differences in hand generation quality across various T2I models. This superior performance at the model level can be attributed to its inherent robustness. By averaging predictions over images generated by the same T2I model, model-level evaluation effectively mitigates the impact of individual errors or outliers, resulting in more stable and consistent assessments. In contrast, image-level evaluation is more sensitive to local prediction noise, which leads to lower performance metrics compared to model-level evaluation.
Additionally, the fine-tuned methods also show improved performance over their original counterparts, validating the practical value of our dataset in supporting hand-aware quality modeling. However, even with such fine-tuning, their performance still lags behind \textit{HandEval}, further highlighting the superiority of our approach in hand quality evaluation. 

Moreover, to demonstrate the effectiveness of \textit{HandEval}, we present several qualitative examples in Fig.~\ref{fig:eval_example}. As shown, our model's scores generally align well with human judgment across a wide range of scenarios,  intuitively demonstrating the strong hand quality perception capability of our model.

\subsubsection{Ablation study on keypoint priors}
To verify the effectiveness of keypoint priors in our hand quality assessment method, we compare the model performance under different settings, as shown in Table~\ref{tab:ablation}. The structure in parentheses indicates the type of keypoint encoder used. The results show that incorporating keypoint priors consistently improves performance across all evaluation metrics. Among them, the model with a GCN-based feature extractor achieves the best performance. This demonstrates that introducing hand keypoint structural information can prominently enhance the model’s ability to assess hand regions. In particular, using graph-based modeling, which is highly aligned with the structural characteristics of hand keypoints, further facilitates the capture of structural dependencies among different hand components, thereby improving the accuracy of quality evaluation.  

\subsubsection{Effect of training data size}
\begin{figure}[htbp]
    \centering
    \includegraphics[width=0.98\linewidth]{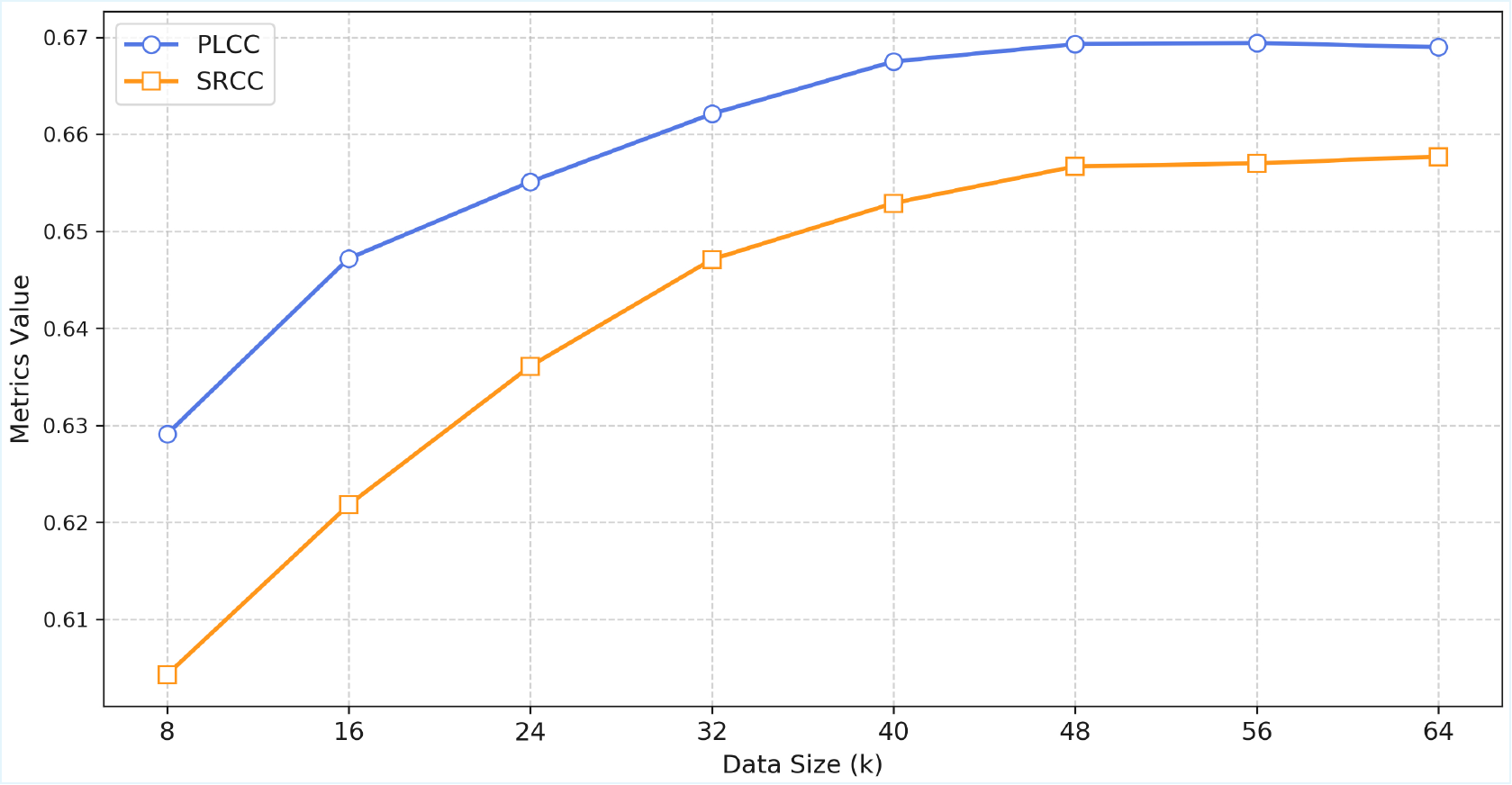}
    \caption{The performance of our model under different training dataset size. We choose 48k as the final dataset size.}
    \label{fig:data_size}
\end{figure}
We also conduct an experiment on the size of training data to verify our dataset scale. The results are shown in Fig.~\ref{fig:data_size}. As observed, the evaluation metrics increase with the growth of the data size, from 8k to 64k. In the beginning, they exhibit a rapid increase, which then slow down, with the values reaching near their maximum at 48k and increasing very slowly thereafter. This verifies our training dataset size at 48K, which servers a good balance for accuracy and training efficiency.
\begin{table}[htbp]
  \centering
  \caption{Performance comparison under different keypoint prior settings.}
  \label{tab:ablation}
  \resizebox{0.45\textwidth}{!}{
  \begin{tabular}{l|c|c|c}
    \toprule
    \multirow{1}{*}{Setting} 
    & \multicolumn{1}{c|}{PLCC} 
    & \multicolumn{1}{c|}{SRCC} 
    & \multicolumn{1}{c}{KRCC} \\
    \midrule
    w/o Keypoint Prior & 0.6330 & 0.6107 & 0.4392 \\
    w Keypoint Prior (MLP) & 0.6539 & 0.6404 & 0.4573 \\
    w Keypoint Prior (GCN) & \textbf{0.6693} & \textbf{0.6567} & \textbf{0.4973} \\
    \bottomrule
  \end{tabular}}
\end{table}

\subsection{Experiments on improving the generation quality of hands}
In this section, we integrate \textit{HandEval} into the Concept Slider framework to guide the model in generating higher quality hands. Following the experimental setup of the Concept Slider for hand refinement in Concept Slider, we fine-tune the SDXL~\cite{sdxl} model using LoRA~\cite{lora}, while keeping \textit{HandEval} frozen during training. We compare the hand generation quality across three models: the original SDXL, the model with the Concept Slider, and the model enhanced with \textit{HandEval}. The detailed results are presented below.
\subsubsection{Quantitative results of hand quatiy}
\begin{figure*}[htbp]
    \centering
    \includegraphics[width=0.98\linewidth]{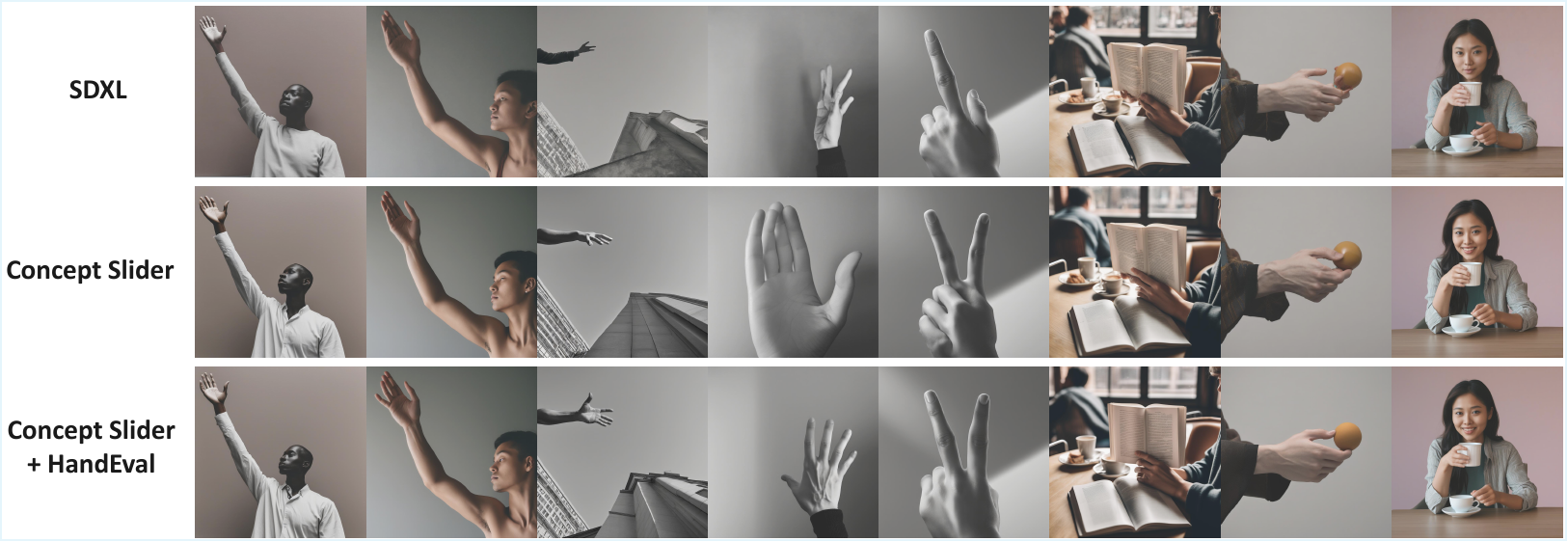}
    \caption{Examples of generated images by SDXL, concept sliders and concept sliders with \textit{HandEval}.}
    \label{fig:generation_example}
\end{figure*}

We generate a total of 120 images using diverse prompts, with 40 images produced by each model. The prompts cover a wide range of hand-related scenarios, including both empty-hand gestures and interactions with objects. The first column of Table~\ref{tab:generation_comparsion} reports the hand quality scores measured by our proposed evaluation model. As shown, Concept Slider, which is fine-tuned based on SDXL, achieves a slight improvement in hand quality over the original SDXL. Notably, when further guided by \textit{HandEval}, the model achieves a substantial enhancement, reaching a score of 3.66. This result indicates that incorporating a hand-specific quality assessment mechanism during generation plays a crucial role in helping the model focus on hand regions and improve local fidelity. Fig.~\ref{fig:generation_example} further illustrates visual comparisons of the generated images by the three models.

\subsubsection{User study} 

We conduct a user study in which participants compare images generated by SDXL, ConceptSlider, and ConceptSlider + \textit{HandEval} under the same prompts, and select the one with the best hand quality. As shown in the second column of Table~\ref{tab:generation_comparsion}, the original SDXL receives only $19.2\%$ of the votes, while ConceptSlider is preferred in $23.9\%$ of cases. In contrast, the model guided by \textit{HandEval} is considered to produce the best hand quality in $56.9\%$ of comparisons. These results further confirm that incorporating hand-specific feedback during generation significantly improves user preference for hand regions in generated images.
\begin{table}[htbp]
  \centering
  \caption{Quantitative comparison and user study of hand quality in generated images.}
  \label{tab:generation_comparsion}
  \resizebox{0.48\textwidth}{!}{
  \begin{tabular}{l|c|c}
    \toprule
    \multirow{1}{*}{Models} 
    & \multicolumn{1}{c|}{Hand Score}
    & \multicolumn{1}{c}{User Preference} \\
    \midrule
    SDXL & 2.31 & 19.2\%\\
    Concept Slider & 2.73 & 23.9\%\\
    Concept Slider + HandEval & \textbf{3.66} & \textbf{56.9\%} \\
    \bottomrule
  \end{tabular}}
\end{table}

\subsection{Experiments on improving AIGC detection performance}
To validate the effectiveness of our proposed hand-enhanced detection method, we construct a small real-world dataset consisting of 200 images for demonstration, each containing visible hand regions. The dataset includes an equal split of $50\%$ real and $50\%$ fake images. Real images are collected from multiple image websites, while fake images are generated using several SOTA T2I models. The reason we don't use existing AIGC detection datasets such as GenImage~\cite{genimage} and WildFake~\cite{wildfake}
is that their human-related data primarily focuses on facial regions, with very few samples containing human hands. We conduct experiments on this dataset to compare the performance of various AIGC detectors with and without our hand-enhanced detection module. The evaluation metrics are $AUC$ (Area Under the ROC Curve), $EER$ (Equal Error Rate), and $Acc$ (Accuracy) at the $EER$ threshold. $AUC$ provides a threshold-agnostic assessment of a model’s overall classification performance. $EER$, defined as the error rate at which the false positive rate equals the false negative rate, also avoids dependence on a specific threshold and reflects a model's discriminative robustness at the most balanced decision boundary. To further assess model performance at this point, we also report $Acc$ at the $EER$ threshold.
\subsubsection{Experiments on different AIGC detectors}

\begin{table*}[htbp]
  \centering
  \caption{aigc detection results enhanced by our hand quality scorer across different detectors. The small font represents the increase/decrease relative to the original detector.}
  \label{tab:detection}
  \resizebox{0.94\textwidth}{!}{
      \begin{tabular}{l|c|c|c|c|c|c}
        \toprule
        \multirow{1}{*}{Method} 
        & \multicolumn{1}{c|}{$AUC$} 
        & \multicolumn{1}{c|}{$EER$} 
        & \multicolumn{1}{c|}{$Acc_{EER}$} 
        & \multicolumn{1}{c|}{$AUC_{refine}$ $\uparrow$}
        & \multicolumn{1}{c|}{$EER_{refine}$ $\downarrow$} 
        & \multicolumn{1}{c}{$Acc^{refine}_{EER}$ $\uparrow$} \\
        \midrule
        UniversalFakeDetect~\cite{universal} & 0.3931 & 0.5941 & 0.4109 & 0.6145\textbf{\tiny +56.3\%} & 0.3366\textbf{\tiny -43.3\%} & 0.6535\textbf{\tiny +59.0\%} \\
        DIRE~\cite{dire} & 0.5200 & 0.4554 & 0.5396 & 0.6410\textbf{\tiny +23.3\%} & 0.3663\textbf{\tiny -19.6\%} & 0.6337\textbf{\tiny +14.8\%} \\
        DRCT~\cite{drct} & 0.7353 & 0.3366 & 0.6584 & 0.7701\textbf{\tiny +4.7\%} & 0.3069\textbf{\tiny -8.8\%} & 0.6881\textbf{\tiny +4.3\%} \\
        DeFake~\cite{defake} & 0.8356 & 0.2475 & 0.7475 & 0.8481\textbf{\tiny +1.5\%} & 0.2376\textbf{\tiny -4.0\%} & 0.7673\textbf{\tiny +2.6\%} \\
        Aeroblade~\cite{aeroblade} & 0.8841 & 0.1782 & 0.8218 & 0.8921\textbf{\tiny +0.9\%} & 0.1683\textbf{\tiny -5.6\%} & 0.8267\textbf{\tiny +0.6\%} \\
        \bottomrule
      \end{tabular}
    }
\end{table*}

\begin{figure*}[htbp]
    \centering
    \includegraphics[width=1.0\linewidth]{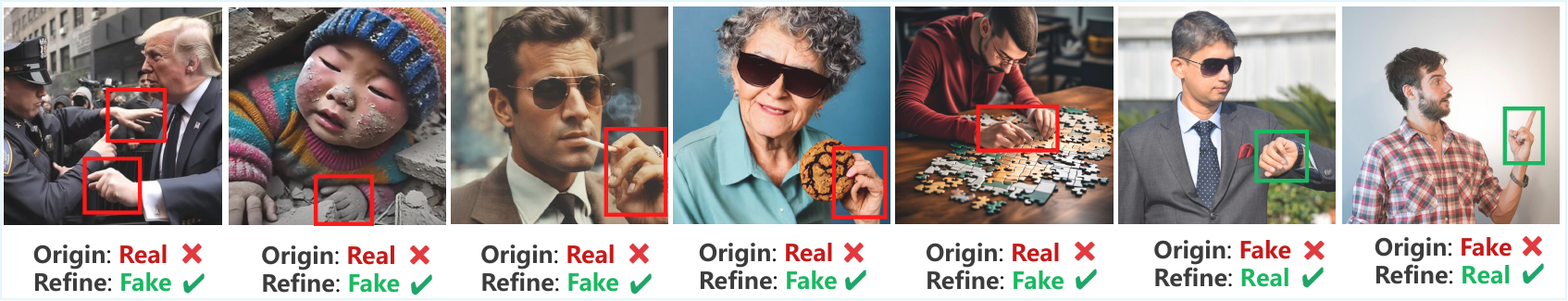}
    \caption{Detection results of the original detectors and those enhanced with hand quality guidance. Leveraging hand quality cues, our method effectively corrects both false positives and false negatives, and can be used to detect AI-generated misinformation images in real-world scenarios (the first two images).}
    \label{fig:detection_refine}
\end{figure*}

Table~\ref{tab:detection} presents the detection performance of the original detectors on our dataset, as well as the results after incorporating hand quality enhancement (denoted with the “refine” suffix). We evaluate five representative detection methods and using their publicly available weights: AEROBLADE~\cite{aeroblade}, which uses autoencoder reconstruction error to detect latent diffusion images; De-Fake~\cite{defake}, which uses image-text classifiers to detect and attribute fakes; DIRE~\cite{dire}, which computes the discrepancy between input and reconstructed images using a pretrained diffusion model; DRCT~\cite{drct}, which creates realistic samples from real images and trains a detector with contrastive learning to better detect fake images from unknown models; and UniversalFakeDetect~\cite{universal}, which performs detection based on nearest-neighbor in CLIP feature space. Note that this evaluation essentially serves as an out-of-distribution cross-dataset test on the data we collected, so it will lead to generally lower performance and also reflects the inherent difficulty of generalization in AIGC detection. 

As can be seen from the results, the introduction of hand quality scores leads to a consistent improvement in detection performance across all models, indicating that hand quality scores can serve as an effective cue for detecting fake images to some extent. It is worth noting that detectors with poorer performance show more significant improvements after incorporating hand quality scores, which is a reasonable phenomenon. This may be because poorer detectors typically rely on coarse and simplified features for judgment, making them prone to overlooking forgery traces in detailed areas such as the hands. Therefore, incorporating hand quality scores can effectively supplement their judgment information, thereby significantly improving their performance. In contrast, high-performance detectors have already considered the hand region to some extent, but may still overlook or misjudge certain details. Therefore, incorporating hand quality scores can still yield a certain performance improvement.

\subsubsection{Experiments on the value of $\alpha$}
\begin{figure}[htbp]
    \centering
    \includegraphics[width=1.0\linewidth]{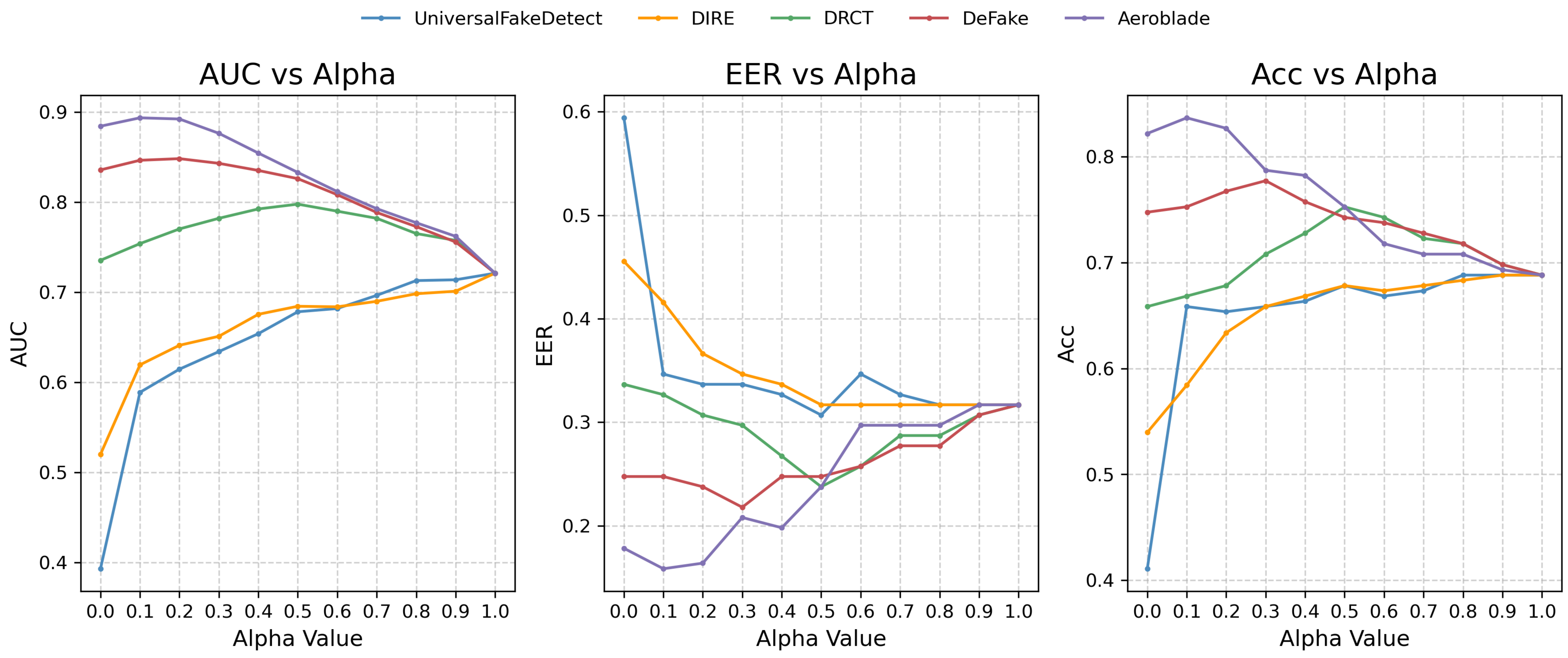}
    \caption{Detection HandEval: Taking the First Step Towards Hand Quality
Evaluation in Generated Imagesperformance under different $\alpha$ settings. A small $\alpha$ values generally benefit stronger detectors like Aeroblade and DeFake, while larger $\alpha$ values are more effective for weaker detectors such as UniversalFakeDetect, DIRE, and DRCT.}
    \label{fig:alpha}
\end{figure}
To further investigate the impact of the weighting parameter $\alpha$ in the fusion strategy between hand quality and detector outputs, we conduct a systematic evaluation of detection performance under different $\alpha$ settings. As illustrated in Fig.~\ref{fig:alpha}, the impact of the $\alpha$ value is closely related to the performance of the detector itself. For stronger detectors, a lower $\alpha$ can enhance detection performance to some extent; whereas for weaker detectors, a larger $\alpha$ can provide a more significant improvement. This phenomenon is reasonable, as mentioned in the previous section: stronger detectors may already consider hand information to a certain extent. Therefore, using a small $\alpha$ as a supplementary signal for hand quality can optimize the detection results without interfering with the original performance. In contrast, weaker detectors are less sensitive to hand regions and require a larger hand quality weight to fully utilize hand region information, compensating for their performance shortcomings. The results presented in the Table~\ref{tab:detection} are under the $\alpha$ value of 0.2, which is a reasonable default based on the performance of all detectors. In practical applications, users can select the most appropriate $\alpha$ value by using a validation set, further optimizing the performance of the detection system.

\subsubsection{Qualitative Results}
Fig.~\ref{fig:detection_refine} intuitively demonstrates the effectiveness of our hand-quality enhancement strategy in the AIGC detection task. The first two images are highly sensationalized generated images in real life (The first image is a fabricated image of Donald Trump arrest incident\footnote{\href{https://www.arabianbusiness.com/industries/technology/donald-trump-arrested-twitter-goes-wild-with-doctored-pictures}{Donald Trump arrested – Twitter goes wild with doctored pictures}}, and the second is a fabricated image of a boy buried under rubble mentioned in the introduction). We compare the results of the original detector with those refined with hand quality guidance. It can be observed that the original detector tends to overlook the quality of the hand region to some extent. In contrast, with our hand-quality-enhanced strategy, the model can effectively reduce both false positives and false negatives by leveraging the quality cues from the hand area. And the detection of the first two real event images shows that our method can leverage hand quality to support AIGC detection in real-world scenarios, helping to prevent potential public controversy.

\section{Conclusion}
In this paper, we address the important issue of poor hand region quality in generated images and the lack of evaluation mechanisms by introducing and systematically studying a novel task: hand quality assessment. We efficiently construct the first dataset for training hand quality assessment models, \textit{HandPair}, which contains 48k hand region images of high and low quality. Based on this dataset, we propose \textit{HandEval}, the first model dedicated to evaluating hand quality. It leverages the powerful detail understanding capabilities of MLLM and further incorporates prior knowledge of hand keypoints, gaining a strong perceptual capability toward hand quality. Experimental results demonstrate that \textit{HandEval} significantly outperforms existing image quality assessment methods in the context of hand region evaluation. Furthermore, we explore the practical value of \textit{HandEval} in two representative downstream tasks. In the image generation task, \textit{HandEval} is integrated into the generation pipeline as a quality supervision signal for hand regions, leading to notable improvements in hand generation quality. In the AIGC detection task, \textit{HandEval} serves as a plug-and-play local quality guidance module, fused with the outputs of existing detectors. This integration consistently improves multiple AIGC detectors' performance, showcasing strong generalization capability and practical utility.

\bibliographystyle{ieeetr}
\bibliography{reference.bib}


\end{document}